\documentclass[conference]{IEEEtran}
\usepackage{blindtext, graphicx}
\usepackage{amsmath}

\begin{document}
%

\title{Representation learning of drug and disease terms for drug repositioning}
\author{\IEEEauthorblockN{Sahil Manchanda}
\IEEEauthorblockA{Department of Computer Science and Engineering\\
Indian Institute of Technology\\
Guwahati, India 781039\\
Email: sahilm1992@gmail.com}
\and
\IEEEauthorblockN{Ashish Anand}
\IEEEauthorblockA{Department of Computer Science and Engineering\\
Indian Institute of Technology\\
Guwahati, India 781039\\
Email: anand.ashish@iitg.ernet.in}}


%


\maketitle
\begin{abstract}
Drug repositioning (DR) refers to identification of novel indications for the approved drugs. The requirement of
huge investment of time as well as money and risk of failure in clinical trials have led to surge
in interest in drug repositioning. DR exploits two major aspects associated with drugs and diseases: 
existence of similarity among drugs and among diseases due to their shared involved genes or pathways or
common biological effects. Existing methods of identifying drug-disease association majorly rely 
on the information available in the structured 
databases only. On the other hand, abundant information available in form of free texts in biomedical 
research articles are not being fully exploited. Word-embedding or obtaining vector representation of words
from a large corpora of free texts using neural network methods have been shown to give significant performance
 for several natural language processing tasks. In this work we propose a novel way of representation
learning to obtain features of drugs and diseases by combining complementary information available in unstructured texts
 and structured datasets. Next we use matrix completion approach on these feature vectors to learn projection matrix between
 drug and disease vector spaces. The proposed method has shown  competitive performance with state-of-the-art methods. Further, the case
 studies on Alzheimer's and Hypertension diseases have shown that the predicted associations are matching with the existing
 knowledge. 
\end{abstract}


%
\IEEEpeerreviewmaketitle

\section{Introduction}

Development of new drugs is associated with huge investment of time and money, and risk of failure in clinical trials.
It has been estimated that on an average, drug development process takes 15 years \cite{dimasi2001new} and associated cost
is approximately \$1 billion \cite{adams2006estimating}.  Finding novel indications for approved drugs, referred as {\it drug repositioning}
or {\it drug repurposing}(DR), has attracted researchers and pharmaceutical industry as a cost-effective and faster alternative to overcome this
 challenge \cite{hurle2013computational}. The candidates for drug repositioning are
drugs which are already in market or which have been discontinued due to various reasons other than safety issues. As per the estimate in
\cite{hurle2013computational}, DR allows a significant reduction in time from 10-17 years to 3-12 years in novel drug discovery.
According to \cite{jin2014toward}, among all the drugs which have been approved by the US Food and Drug Administration (FDA), approximately 
$30\%$ of them were the result of drug repositioning. Significant examples of drug repositioning includes {\it Aspirin} (regular use as analgesic
 and now also being widely adapted to treat heart related disease \cite{wolff2009aspirin}), {\it Plerixafor} (initially developed to treat HIV but
later being used as a drug to mobilize stem cells~\cite{flomenberg2005use}), and {\it Thalidomide} (initially developed to treat nausea but
after drug repositioning research, being used to treat dermatological issues and the myelome disease~\cite{calabrese2000thalidomide}).

There have been significant number of methods developed for the drug repositioning problem including machine learning methods. We summarize the
prominent methods in the section~\ref{sec:relwork}. Working principle of all methods rely on two important aspects related to drugs and diseases.
First, drugs often bind to multiple targets resulting into various biological effects including side-effects~\cite{pujol2010unveiling}. Second, a 
biological target of a drug which is relevant to a particular disease, may also be directly or indirectly associated with other diseases. In other
words, overlapping pathways or common associated targets between various diseases are important factors and thereby making it possible that an approved
drug for one disease may be useful in treating a similar disease~\cite{sardana2011drug}.

Existing methods of identifying drug-disease association majorly rely on the information available in
the structured databases only. However these databases are unable to keep pace with the exponential growth of information
appearing in research articles. In this paper, our primary aim is to develop a method which can exploit information present in free texts as well
as in structured databases. In recent years, vector representation of words, learned using neural network based methods from a large corpora of free
texts, have been shown to give significant performance for several natural language processing tasks. Word vectors thus obtained are also shown to capture
syntactic and semantic properties. We employ a novel method to learn representation of each drug and disease terms which then are projected on a common vector
space to obtain similarity between drugs and diseases. Towards this end, first we learn vector representation of drugs and diseases by using the knowledge 
present in literature. Next, these vectors are updated to accommodate various similarity measures of drugs and diseases respectively. The resultant drug and disease vector representation are not necessarily in the same vector-space. So we employ matrix completion approach~\cite{yu2014largescal} to learn a projection matrix
between drug and disease vector space. We evaluate the performance of our method using ten fold cross validation and top k rank threshold methods and compare it with 3 other competitive methods. We further perform case studies on Alzheimer's disease and Hypertension and verify our predictions for these diseases from literature. Our study shows that all our top ten drugs predicted for Alzheimer's disease are approved to treat neurodegenerative diseases. Similarly 7 out 10 drugs predicted for Hypertension are approved and 2 out of remaining 3 are used to treat Ocular Hypertension.

\section{Materials and Methods}
In this study we describe the datasets used. Later we explain our method which includes learning the feature vector and learning the projection matrix between drug and disease vector space. 

\subsection{Dataset}
\label{sec:dataset}
This section discusses all datasets and their sources used in this work. 

\begin{enumerate}
 \item {\bf Drug-Disease Association Data-} 

We use the same drug-disease association data as used in PREDICT \cite{gottlieb2011predict}. Data is made available as the
supplementary material of the corresponding paper \cite{gottlieb2011predict}. It contains $1933$ drug-disease associations between
$593$ drugs and $313$ diseases. All $593$ drugs are registered in DrugBank \cite{DrugBank} and all $313$ diseases are listed in 
OMIM \cite{Omim} database. As we have mentioned earlier that the proposed method rely on obtaining word embedding 
for each drug and disease from
a huge corpus of biomedical articles, we discarded some drugs and diseases which we were not present in the corpus. 
Finally we consider $584$ drugs, $294$ diseases and $1854$ drug-disease associations known between them.

 \item {\bf SIDER: Drug Side Effect Data-} We obtain the list of side effects corresponding to all $584$ drugs from the SIDER~\cite{sider} database.
 \item {\bf Chemical Fingerprint of Drugs-}	Chemical fingerprint of a drug corresponds to the record of component fragment present in their chemical structure. For each
drug their chemical fingerprint was obtained from the DrugBank~\cite{DrugBank}.
 \item {\bf DrugBank-} Similar to the chemical fingerprint of drugs, drug targets are obtained from the DrugBank. Drug targets one or more cellular molecules such as metabolites or proteins for desired effects. A list of targets corresponding
to all $584$ drugs are obtained.
 \item {\bf DisGeNet: Disease Associated Genes-} We collect genes associated with disease from DisGeNET ~\cite{pinero2015disgenet}.
\end{enumerate}

\subsection{Construction of similarity measures}
We calculate three types of similarity for drugs which are based upon side-effects, chemical structure and target proteins. Two similarities are calculated for each disease pair based upon the disease phenotypes and associated genes.

\subsubsection{Drug Similarity measures}

\begin{enumerate}
  \item[I] Side effect similarity : A side effect is an undesired consequence of a drug.  Drugs cause side-effects when they bind to
off-target  apart  from  their  desired  on-targets. Under  the  assumption  that  if  2  drugs  share
side-effects  and  hence  off-targets,  there  is  a  possibility that  they  might  share  on-targets  which
can be used to cure diseases. Studies \cite{campillos2008drug} show that drugs sharing off
targets might also share on targets. For each drug pair ($d_i,d_j$), this similarity is :\newline
             \[ Sim(d_i,d_j) = \frac{|SE(d_i) \cap SE(d_j)|}{|SE(d_i) \cup SE(d_j)|}  \]
             where $SE(d)$ is the set of side-effects related to drug $d$.

\item[II] Chemical Similarity : Similarity of two chemicals is based upon comparing their chemical  fingerprint.  A  fingerprint
is a record of component fragment present in a chemical structure. It has been shown in \cite{bajusz2015tanimoto} that {\it Tanimoto coefficient} can be an effective measure to 
calculate similarity between two chemicals based on their structures. Pairwise similarity between two drugs was calculated as Tanimoto score of their fingerprint using RDKit \cite{Rdkit} library of Python.

\item[III] Drug-Target Similarity : A biological target is the protein in the body which is either up regulated or down regulated due
to the action of a particular drug on it .
If two drugs share same targets, the probability of them causing the similar effect may also increase. Pairwise drug-target similarity between drugs  $d_i$ and $d_j$  is calculated as :

\begingroup\makeatletter\def\f@size{8}\check@mathfonts
\def\maketag@@@#1{\hbox{\m@th\large\normalfont#1}}%
\begin{align*}
    Sim(d_i,d_j)=\frac{1}{|P(d_i)||P(d_j)| } \sum_{i=1}^{P(d_i)}\sum_{j=1}^{P(d_j)} SW(P(d_i),P(d_j)) 
\end{align*}\endgroup
    where $P(d)$ denotes the  set of genes associated to drug $d$ and $SW$ is the Smith-Waterman Sequence alignment score \cite{smith1985statistical}.

\end{enumerate}

\subsubsection{Disease Similarity measures}

\begin{enumerate}
   \item[I] Phenotypic similarity : A phenotypic feature is an observable biological or clinical characteristic of a disease. It is a amalgamation of gene
expression as well as influence of external environmental factors. The similarity is collected from MIMMiner Tool ~\cite{van2006text}. The tool measures  disease similarity by computing similarity between MeSH terms \cite{lipscomb2000medical} that appear  in  the  medical  description  of  diseases  in  the OMIM database.

   \item[II] Gene Similarity : Disease causing or associated genes are collected from DisGeNET \cite{pinero2015disgenet}. Pairwise gene similarity between disease $d_i$ and $d_j$ is calculated as:
 \begingroup\makeatletter\def\f@size{8}\check@mathfonts
\def\maketag@@@#1{\hbox{\m@th\large\normalfont#1}}%
\begin{align*}
    Sim(d_i,d_j) = \frac{1}{|P(d_i)||P(d_j)| }\sum_{i=1}^{P(d_i)}\sum_{j=1}^{P(d_j)}SW(P(d_i),P(d_j))
\end{align*}\endgroup

    where $P(d)$ denotes the set of genes associated to disease $d$ and $SW$ is the Smith-Waterman Sequence alignment score \cite{smith1985statistical}.
\end{enumerate}

\subsection{Method}
\label{sec:method}
\begin{figure}
  \includegraphics[scale=0.26]{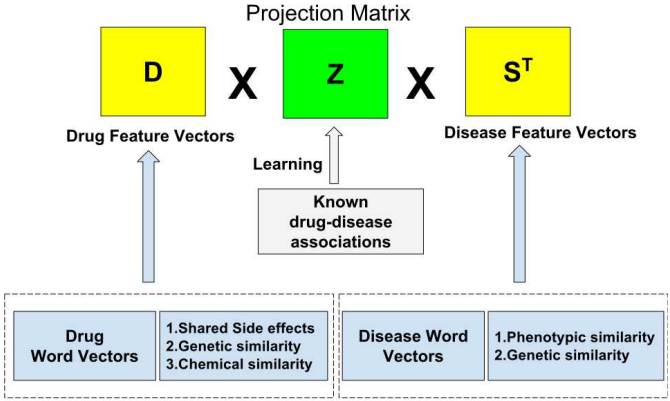}
    \caption{Diagram depicting the flow of our method. First, the drug and disease feature vector are learned. To do this, drug word vector and disease word vector are updated using the similarity measures. Second, the projection matrix is learned by using the well known associations and our drug/disease feature vectors. }
    \label{fig:method}
\end{figure}

The proposed method has three major steps. In the first step we obtain vector representation of
drugs and diseases using neural embedding method \cite{mikolov2013distributed}. We update these representations using similarity
scores calculated from the various structured datasets. And in the last step, we learn a projection
matrix between the two vector-spaces so that a final association score between drug-disease pair
can be obtained. It is noteworthy to mention here again that there is no requirement of negative
datasets. Figure~\ref{fig:method} summarizes the proposed method.

\subsubsection{Word vectors for drugs and diseases}
To capture the information present in literature, we obtain the word vector representation of drugs and diseases. We use Pubmed \cite{pubmed} open access set as our corpus. Each disease
is mapped to its OMIM id. As diseases can appear under various names in Pubmed \cite{pubmed} corpus, each disease in the corpus is mapped to a Concept Unique Identifier(CUI) by using UMLS Meta thesaurus \cite{umls}. If a disease (OMIM indication) has multiple concept names associated to it, then the resultant vector is taken as the simple average of all the vectors associated to that OMIM indication. The concept names for each OMIM indication is obtained from Supplementary Information of PREDICT \cite{gottlieb2011predict}.  Word vector representation of each drug and disease is obtained by training Pubmed Corpus using word2Vec \cite{word2vec} Python library. To train vectors, we set window size to 5. We have experimented using various vector dimensions ranging from 100 to 200.  

\subsubsection{Learning vector representation by combining similarity measures}
Let $N_d$ be the number of drugs and $N_s$ be the number of diseases.
Each drug word vector is denoted as $d_i \in \mathcal{R}^N$, where $i$ ranges from $1$ to $N_d$.
Each disease word vector is denoted as $s_i \in \mathcal{R}^N$, where $i$ ranges from $1$ to $N_s$.
Let the updated drug vector (feature vector) for i\textsuperscript{th} 
drug be denoted as $\tilde{d_i}$, which is initialized to $d_i$.
Let the updated disease vector (feature vector) for i\textsuperscript{th} disease be denoted as $\tilde{s_i}$, which is initialized to $s_i$. Let $Sim_k(i,j)$ denote the k\textsuperscript{th} similarity between drug i and drug j or disease i and disease j.

Let $M$ be the number of drug similarity measures and $L$ be the number of disease similarity measures.
The motive is to obtain a feature vector for each drug and disease by combining the above mentioned similarities and updating the word vectors. For each drug $i$ ( $i$ varies from 1 to $N_d$), $\tilde{d_i}$ is updated when the below objective ( $J_1$ ) is minimized:
\[J_1=\sum_{j=1}^{N_d}\sum_{k=1}^{M}(\frac{\tilde{d_i} . d_j} { |\tilde{d_i}||d_j|  } - Sim_k(i,j))^2\]
where $|d_i|$ denote the length of the vector $d_i$  .\newline

Each drug word vector is updated using all the other drug vectors and for each similarity measure. The updated set of drug vectors (called feature vectors) is denoted as $D=[\tilde{d_1},\tilde{d_2},...,\tilde{d_{N_d}}] $, where each $\tilde{d_i}\in \mathcal{R}^N$.
\newline

Similar kind of objective ( $J_2$ ) is minimized for all disease $\tilde{s_i}$, where $i$ varies from 1 to $N_s$ .
\[J_2=\sum_{j=1}^{N_s}\sum_{k=1}^{L}(\frac{\tilde{s_i} . s_j} { |\tilde{s_i}||s_j|  } - Sim_k(i,j))^2\]

where $|s_i|$ denote the length of the vector $s_i$.\newline

Each disease word vector is updated using all the other disease vectors and for each similarity measure. The updated set of disease vectors (called feature vectors) is denoted as $S=[\tilde{s_1},\tilde{s_2},...,\tilde{s_{N_s}} ] $, where each $\tilde{s_i}\in \mathcal{R}^N$.

The optimization problem is solved using Theano \cite{theano} library of Python. We have obtained a drug vector space and a disease vector space where each vector is of dimension $\mathcal{R}^N$.

\subsubsection{Learning projection from drug vector space to disease vector space}
 Our motive is to learn a projection matrix from drug vector space to disease vector space which will help us in predicting drug-disease association scores. The projection  matrix should be such that the projected drug vectors are geometrically close to vectors of their well known  disease vectors. The drugs that are in proximity in the directions of their feature vectors may share diseases and vice-versa.
Let $I \in \mathcal{R}^{N_d \times N_s}$ be called the association matrix
where $I_{ij}$ is 1 if $drug_i$ treats $disease_j$ else 0.
The projection matrix is denoted as Z $\in\mathcal{R}^{N\times N}$ .

To learn this projection matrix we use inductive matrix completion approach \cite{yu2014largescal}\cite{ natarajan2014inductive} which minimizes the following objective function: 
    \[ \min\limits_{G,H} \sum_{i,j} ||I_{ij} - \tilde{d_i}GH^T \tilde{s_j}^ T ||^2 + \frac{\lambda}{2} ( ||G||^2 + ||H||^2 )\]    
    where the projection matrix $Z = GH^T$, where $G\in \mathcal{R}^{N \times K}$ and  $H \in \mathcal{R}^{N \times K}$.
    The score of a drug $i$ and disease $j$ pair is calculated as:
        \[score(i,j) = \tilde{d_i}Z\tilde{s_j}^T\]
Higher the score, greater is the possibility of drug i treating disease j .    

\section{Experiments}


We conduct 10-fold cross-validation experiments to evaluate the performance of all methods. We use
AUC, ROC and top-rank thresholds as evaluation metrics. In the top-rank threshold measure, a well
known drug-disease association is considered as correctly predicted if its rank based on the predicted
score is within the specified rank threshold.

\subsection{Baseline Methods}
We compare the proposed method with three other methods, MBIRW~\cite{luo2016drug}, HGBI~\cite{wang2013drugHET}
and TP-NRWRH~\cite{liu2016inferring}. We briefly summarize each of the three methods 
for the sake of completeness. 

The HGBI method creates a heterogeneous network of two different type of nodes. One set of nodes are representing different drugs
and another set of nodes represent targets. Edges exist between within same node types as well as between two different node types.
Existence of edge depends on drug-drug similarities, target-target similarities and drug-target interactions. The edge weights of
the network are updated in an iterative fashion by incorporating all the paths between the drug-target pair.

The MBIRW method constructs two separate networks on drugs and diseases. Both similarity networks were created using novel 
similarity measure which takes into account correlation between different similarities. Further MBIRW performs bi-directional
random walk on these two networks to get scores for drug disease associations.

TP-NRWH again uses random walk method but on single heterogeneous drug-disease network. This network is similar to the network
used in the the HGBI method and integrates all the similarity measures (drug-drug and disease-disease) and well known
drug-disease associations. This is in contrast to the MBIRW method which creates two separate networks on drugs and diseases.

We use default parameter settings for all the three methods. Parameters of TP-NRWRH are 
set as ($\alpha=0.3,\lambda=0.8,\eta=0.4$). For MBIRW $\alpha$ is set to default 0.3 and max iterations for right and left random walk is set to 2. For HGBI, restart probability $\alpha$ is set to default 0.4 and cut off was set to 0.3.

\section{Results}

\subsection{Vector representation}
\begin{figure}
  \includegraphics[scale=0.55]{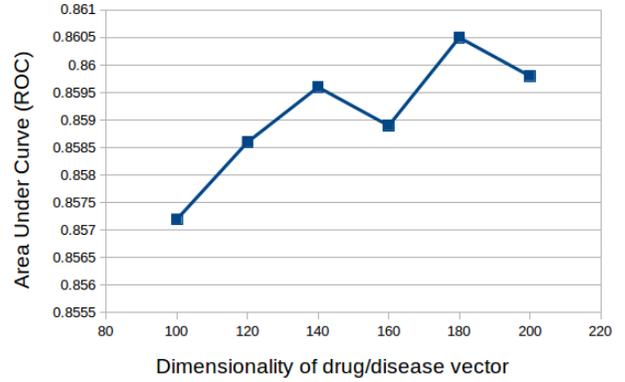}
    \caption{Performance of our method in terms of AUC with respect to different dimensionality of feature vector of drugs and diseases. }
    \label{fig:dim_vector}
\end{figure}

First we analyze the performance of our method with respect to varying length of feature vectors between 100 and 200. 
Fig~\ref{fig:dim_vector} shows AUC obtained by using different size of drugs and disease vectors. Although increasing
 dimension generally led to improved AUC score but improvement was not really significant. Next we analyze the importance
of updating the word vectors based on similarity scores. We obtain an AUC score of $0.77$ when word-vectors obtained using
word2vec method on biomedical copora are not updated. On the other hand an  AUC score of $0.86$ is obtained when updated
word-vectors are used. The relative improvement of $10\%$ clearly indicates that the vectors learned through our method
captured the similarity of drugs and diseases in better manner.

\subsection{Comparison with existing methods}
\begin{figure}[hbt]
  \includegraphics[scale=0.55]{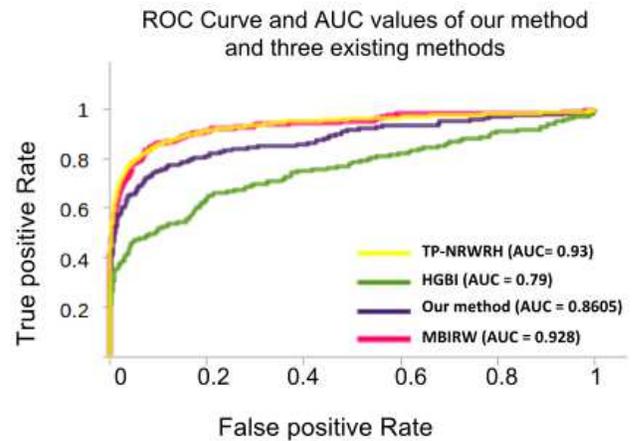}
    \caption{Diagram depicting the Receiver operating characteristics of our method and 3 other competitive methods. The AUC values are also mentioned.}
    \label{fig:roc_comparison}
\end{figure}

\begin{figure}
  \includegraphics[scale=0.27]{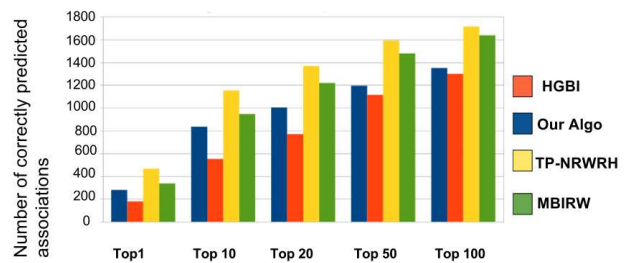}
    \caption{Diagram depicting the number of correctly predicted associations with respect to five different top-ranked thresholds.}
    \label{fig:toprank}
\end{figure}

Fig~\ref{fig:roc_comparison} shows the AUC values and the ROC of 10-fold cross-validation experiments. Although the proposed method 
has obtained AUC value of $0.86$ which is better than the one obtained by HGBI ($0.79$) but the other two methods were 
best performing methods.
Similar observations are made based on the top-rank threshold metric. The number of correctly predicted
associations by our method is greater than that of HGBI for every top rank thresholds as shown in the Fig~\ref{fig:toprank}.

\subsection{Case Studies}
After finding the performance of our model, we conducted leave-disease-out experiment. For this, first we select a disease and train our model only with the 
remaining data after excluding all known associations related to it. Then scores are calculated for the held out disease and top scoring drugs are reported.
We perform the case studies on two diseases, Alzheimer and Hypertension.

\begin{enumerate}
 \item Alzheimer's Disease: Table~\ref{tab:alzheimer} shows the top scoring drugs predicted by the model for Alzheimer's disease.
Out of the top 10 drugs predicted by our method, 6 drugs namely {\it Rivastigmine}, {\it Galantamine}, {\it Donepezil}, 
{\it Memantine}, {\it Tacrine} and {\it Valproic Acid} have been approved for Alzheimer's disease. Other drugs namely 
{\it Ropinirole}, {\it Entacapone}, {\it Pramipexole} and {\it Carbidopa} have been used to treat Parkinson's disease.  
Although there is a difference in pathogenesis of Parkinson's disease and Alzheimer's disease, but both of them are
neuro-degenerative disease associated with aging \cite{nussbaum2003alzheimer}. Pramipexole has been under Phase 2 of Clinical trials (NCT01388478)~\cite{ClinicalTrials} for the treatment of Alzheimer's disease. 
The above results have been verified from PREDICT \cite{gottlieb2011predict} and DrugBank \cite{DrugBank}. 

\begin{table}[]
\centering
\caption{TOP 10 PREDICTED DRUGS FOR ALZHEIMER'S DISEASE BY
OUR METHOD}
\label{tab:alzheimer}
\begin{tabular}{ccccc}
Rank                     & Drug Name                          & \begin{tabular}[c]{@{}c@{}}Predicted \\ Score\end{tabular} & \begin{tabular}[c]{@{}c@{}}Clinical \\ Evidence\end{tabular} & Mean Score                    \\ \hline
\multicolumn{1}{|c|}{1}  & \multicolumn{1}{c|}{Rivastigmine}  & \multicolumn{1}{c|}{0.31896}                               & \multicolumn{1}{c|}{Yes}                                     & \multicolumn{1}{c|}{0.00090}  \\ \hline
\multicolumn{1}{|c|}{2}  & \multicolumn{1}{c|}{Galantamine}   & \multicolumn{1}{c|}{0.21839}                               & \multicolumn{1}{c|}{Yes}                                     & \multicolumn{1}{c|}{0.00274}  \\ \hline
\multicolumn{1}{|c|}{3}  & \multicolumn{1}{c|}{Donepezil}     & \multicolumn{1}{c|}{0.20712}                               & \multicolumn{1}{c|}{Yes}                                     & \multicolumn{1}{c|}{0.00669}  \\ \hline
\multicolumn{1}{|c|}{4}  & \multicolumn{1}{c|}{Memantine}     & \multicolumn{1}{c|}{0.19311}                               & \multicolumn{1}{c|}{Yes}                                     & \multicolumn{1}{c|}{0.00671}  \\ \hline
\multicolumn{1}{|c|}{5}  & \multicolumn{1}{c|}{Ropinirole}    & \multicolumn{1}{c|}{0.15378}                               & \multicolumn{1}{c|}{No}                                      & \multicolumn{1}{c|}{0.01309}  \\ \hline
\multicolumn{1}{|c|}{6}  & \multicolumn{1}{c|}{Tacrine}       & \multicolumn{1}{c|}{0.14735}                               & \multicolumn{1}{c|}{Yes}                                     & \multicolumn{1}{c|}{0.00309}  \\ \hline
\multicolumn{1}{|c|}{7}  & \multicolumn{1}{c|}{Entacapone}    & \multicolumn{1}{c|}{0.14210}                               & \multicolumn{1}{c|}{No}                                      & \multicolumn{1}{c|}{0.00625}  \\ \hline
\multicolumn{1}{|c|}{8}  & \multicolumn{1}{c|}{Valproic Acid} & \multicolumn{1}{c|}{0.14069}                               & \multicolumn{1}{c|}{Yes}                                     & \multicolumn{1}{c|}{0.01276}  \\ \hline
\multicolumn{1}{|c|}{9}  & \multicolumn{1}{c|}{Pramipexole}   & \multicolumn{1}{c|}{0.12957}                               & \multicolumn{1}{c|}{No}                                      & \multicolumn{1}{c|}{-0.00299} \\ \hline
\multicolumn{1}{|c|}{10} & \multicolumn{1}{c|}{Carbidopa}     & \multicolumn{1}{c|}{0.12407}                               & \multicolumn{1}{c|}{No}                                      & \multicolumn{1}{c|}{0.00282}  \\ \hline
\end{tabular}
\end{table}

 \item Hypertension: Another indication, Hypertension was analysed and its top scoring drugs are shown in Table~\ref{tab:hypertension}. Out of the top 10 drugs predicted by our method, 7 drugs namely {\it Trandolapril}, {\it Prazosin}, {\it Mecamylamine}, {\it Labetalol}, {\it Captopril}, {\it Losartan} and 
 {\it Valsartan} are approved drugs for hypertension. {\it Brinzolamide} and {\it Travoprost} have been used for treating ocular hypertension. 
The scores of these top predicted drugs for treating Hypertension and  Alzheimer's disease are much higher than the mean score of these drugs for all diseases.
\begin{table}[]
\centering
\caption{TOP 10 PREDICTED DRUGS FOR HYPERTENSION BY OUR
METHOD}
\label{tab:hypertension}
\begin{tabular}{ccclc}
Rank                     & Drug Name                         & \begin{tabular}[c]{@{}c@{}}Predicted \\ Score\end{tabular} & \begin{tabular}[c]{@{}l@{}}Clinical \\ Evidence\end{tabular} & Mean Score                   \\ \hline
\multicolumn{1}{|c|}{1}  & \multicolumn{1}{c|}{Dyphylline}   & \multicolumn{1}{c|}{0.33169}                               & \multicolumn{1}{l|}{No}                                      & \multicolumn{1}{c|}{0.01341} \\ \hline
\multicolumn{1}{|c|}{2}  & \multicolumn{1}{c|}{Trandolapril} & \multicolumn{1}{c|}{0.30106}                               & \multicolumn{1}{l|}{Yes}                                     & \multicolumn{1}{c|}{0.01325} \\ \hline
\multicolumn{1}{|c|}{3}  & \multicolumn{1}{c|}{Prazosin}     & \multicolumn{1}{c|}{0.27290}                               & \multicolumn{1}{l|}{Yes}                                     & \multicolumn{1}{c|}{0.01150} \\ \hline
\multicolumn{1}{|c|}{4}  & \multicolumn{1}{c|}{Mecamylamine} & \multicolumn{1}{c|}{0.26114}                               & \multicolumn{1}{l|}{Yes}                                     & \multicolumn{1}{c|}{0.01084} \\ \hline
\multicolumn{1}{|c|}{5}  & \multicolumn{1}{c|}{Brinzolamide} & \multicolumn{1}{c|}{0.26103}                               & \multicolumn{1}{l|}{No}                                      & \multicolumn{1}{c|}{0.01857} \\ \hline
\multicolumn{1}{|c|}{6}  & \multicolumn{1}{c|}{Travoprost}   & \multicolumn{1}{c|}{0.25957}                               & \multicolumn{1}{l|}{No}                                      & \multicolumn{1}{c|}{0.00937} \\ \hline
\multicolumn{1}{|c|}{7}  & \multicolumn{1}{c|}{Labetalol}    & \multicolumn{1}{c|}{0.25421}                               & \multicolumn{1}{l|}{Yes}                                     & \multicolumn{1}{c|}{0.01268} \\ \hline
\multicolumn{1}{|c|}{8}  & \multicolumn{1}{c|}{Captopril}    & \multicolumn{1}{c|}{0.25322}                               & \multicolumn{1}{l|}{Yes}                                     & \multicolumn{1}{c|}{0.01211} \\ \hline
\multicolumn{1}{|c|}{9}  & \multicolumn{1}{c|}{Losartan}     & \multicolumn{1}{c|}{0.25189}                               & \multicolumn{1}{l|}{Yes}                                     & \multicolumn{1}{c|}{0.01750} \\ \hline
\multicolumn{1}{|c|}{10} & \multicolumn{1}{c|}{Valsartan}    & \multicolumn{1}{c|}{0.24950}                               & \multicolumn{1}{l|}{Yes}                                     & \multicolumn{1}{c|}{0.00656} \\ \hline
\end{tabular}
\end{table}

\end{enumerate}

\section{Related work}
\label{sec:relwork}

Several computational methods have been developed to solve the drug repositioning problem including machine learning as well as literature mining based methods.
Literature mining methods mainly rely on co-occurrence of drug, disease and targets within a context ~\cite{andronis2011literature}\cite{napolitano2013drug}. 
These methods generally have poor performance as they do not use any contextual semantic information and treat all types of relations between two relevant terms
equivalently. On the other hand, machine learning based models have been shown to perform relatively better than co-occurrence based methods. 
Gottlieb et al. \cite{gottlieb2011predict} developed a logistic regression based model and combined various similarity measures of drugs and diseases in order to predict drug-disease associations. Wang et al. \cite{wang2013drug} applied Support Vector Machine model on drug chemical structures, protein sequences and disease phenotypic data in order to identify new relations between drugs and diseases. One of the major issue faced by these methods is requirement of negative data, which
is not available.

Network based models are good alternative as they do not require both positive and negatively labeled data.
Chiang et al. developed a network based model which predict novel indications for drugs based upon the fact that if two disease share treatment profiles, then drugs approved for one of those disease may be used to treat the other disease too. The method relied upon "guilt by association" technique \cite{chiang2009systematic}.  Wang et al. developed a novel method which consists of three-layer heterogeneous graph model to integrate relationships between drugs, diseases and targets. Based on this model, an iterative algorithm  was  used in order to  rank drugs  for  each disease. The method measures the strength between unlinked drug-target pairs by using all the paths in the network \cite{wang2014drug}. Martinez et al. developed the DrugNet method, a network -based prioritization method, that utilizes the information of drugs, diseases and drug targets and performs prioritization of drug-disease and disease-drug  \cite{martinez2015drugnet} associations. Lee et al. exploited the structural properties of biological networks. They developed shared neighborhood scoring algorithm and applied it on an integrated drug-protein-disease tripartite network in order to predict new indications for drugs \cite{lee2012rational}. Chen et al. utilized already existing recommendation systems ProbS and Heats to give recommendation score for disease with respect to drugs \cite{chen2015network}. Wu et al. constructed a weighted heterogeneous network of drugs and diseases and applied clustering technique using ClusterONE \cite{nepusz2012detecting} algorithm to predict new drug-disease associations \cite{wu2013computational}. Luo et al. improved drug-drug and disease-disease similarity measures by using concepts of clustering and by performing various analysis on similarity values. Further they performed a bi-random walk on the similarity network and well known drug-disease associations and learned drug-disease association matrix \cite{luo2016drug}.

\section{Conclusion}
In this paper we have presented a novel representation learning method to obtain vector representation of drugs and diseases. These representations are then utilized
to obtain association score between drug-disease pairs.
The main contribution of this work is combining complementary information available in unstructured texts and structured datasets. Heterogeneous information was combined and feature vectors were learned for drugs and diseases. Prediction using updated feature vectors gave better results than using the original word vectors.
Case studies on Hypertension and Alzheimer's disease indicate that predictions made by our method can be used for biomedical research. We compared our method with existing methods on drug repositioning. Our results are fairly comparable to those methods in terms of AUC and top k rank threshold scoring mechanism.


%

\section*{Acknowledgment}

The authors would like to thank Shubhakar Reddy (Former bachelor's student at Indian Institute of Technology Guwahati) for providing us the disease to CUI mapping tool developed by him.

\ifCLASSOPTIONcaptionsoff
  \newpage
\fi



\bibliographystyle{IEEEtran}
\bibliography{references}
%



%

\begin{IEEEbiography}[{\includegraphics[width=1in,height=1.25in,clip,keepaspectratio]{picture}}]{John Doe}
\blindtext
\end{IEEEbiography}




\end{document}